\documentclass[letterpaper]{article} 
\usepackage{aaai25}  
\usepackage{times}  
\usepackage{helvet}  
\usepackage{courier}  
\usepackage[hyphens]{url}  
 \usepackage{booktabs} 
\usepackage{graphicx} 
\usepackage{natbib}  
\usepackage{caption} 
\usepackage{algorithm}
\usepackage{algorithmic}
\usepackage{comment}
\usepackage{graphicx}
\usepackage{latexsym}

\usepackage{amsmath}
\usepackage{amssymb}
\usepackage{mathtools}
\usepackage{tikz}
\usetikzlibrary{positioning, arrows}

\usepackage{subcaption}
\usepackage{newfloat}
\usepackage{listings}
\DeclareCaptionStyle{ruled}{labelfont=normalfont,labelsep=colon,strut=off} 
\floatstyle{ruled}
\newfloat{listing}{tb}{lst}{}
\floatname{listing}{Listing}
\title{Autonomous Goal Detection and Cessation in Reinforcement Learning: A Case Study on Source Term Estimation}
\author{Yiwei Shi\textsuperscript{\rm 1} \thanks{This work is sponsored by University of Bristol Scholarship.}, Muning Wen\textsuperscript{\rm 2}, 
Qi Zhang\textsuperscript{\rm 3}, 
Weinan Zhang\textsuperscript{\rm 2}, 
Cunjia Liu\textsuperscript{\rm 4}, 
Weiru Liu\textsuperscript{\rm 1}}

\affiliations{\textsuperscript{\rm 1} School of Engineering Mathematics and Technology, University of Bristol\\\textsuperscript{\rm 2}  Department of Computer Science \& Engineering, Shanghai Jiao Tong University\\
\textsuperscript{\rm 3}  School of Electronic and Information Engineering, Tongji University\\
\textsuperscript{\rm 4}  Department of Aeronautical and Automotive Engineering, Loughborough University\\
yiwei.shi@bristol.ac.uk}

\usepackage{bibentry}

\begin{document}

\maketitle

\begin{abstract}
Reinforcement Learning has revolutionized decision-making processes in dynamic environments, yet it often struggles with autonomously detecting and achieving goals without clear feedback signals. For example, in a Source Term Estimation problem, the lack of precise environmental information makes it challenging to provide clear feedback signals and to define and evaluate how the source's location is determined. To address this challenge, the Autonomous Goal Detection and Cessation (AGDC) module was developed, enhancing various RL algorithms by incorporating a self-feedback mechanism for autonomous goal detection and cessation upon task completion. Our method effectively identifies and ceases undefined goals by approximating the agent's belief, significantly enhancing the capabilities of RL algorithms in environments with limited feedback. To validate  effectiveness of our approach, we integrated AGDC with deep Q-Network, proximal policy optimization, and deep deterministic policy gradient algorithms, and evaluated its performance on the Source Term Estimation problem. The experimental results showed that AGDC-enhanced RL algorithms significantly outperformed traditional statistical methods such as infotaxis, entrotaxis, and dual control for exploitation and exploration, as well as a non-statistical random action selection method. These improvements were evident in terms of success rate, mean traveled distance, and search time, highlighting AGDC's effectiveness and efficiency in complex, real-world scenarios.
\end{abstract}

%

\section{Introduction} \label{sec:Introduction}

Reinforcement Learning (RL) optimizes decision-making and behavior in dynamic environments through trial and error and reward mechanisms. It is widely used in fields such as gaming \cite{mnih2015human,silver2017mastering,wang2024zsc,wang2022model}, robotic control \cite{levine2016end,zhang2021model,wang2023order}, autonomous driving \cite{feng2023dense,kiran2021deep}, industrial automation \cite{degrave2022magnetic,meyes2017motion}, preference alignment \cite{NEURIPS2022_8be9c134,Bai_Zhang_Tao_Wu_Wang_Xu_2023,bai2024efficient} and collaborative tasks \cite{li2023cooperative,li2024tackling,xu2024beyond}. RL involves agents interacting with the environment, learning through exploration and exploitation to achieve their goals. However, in real-world applications like the Source Term Estimation (STE) problem \cite{steiner2001large}, the environment often lacks clear signals for the end of an episode or direct rewards, making it difficult for the agent to determine task completion.

\begin{figure}[!tb]
    \centering
    \includegraphics[width=\linewidth]{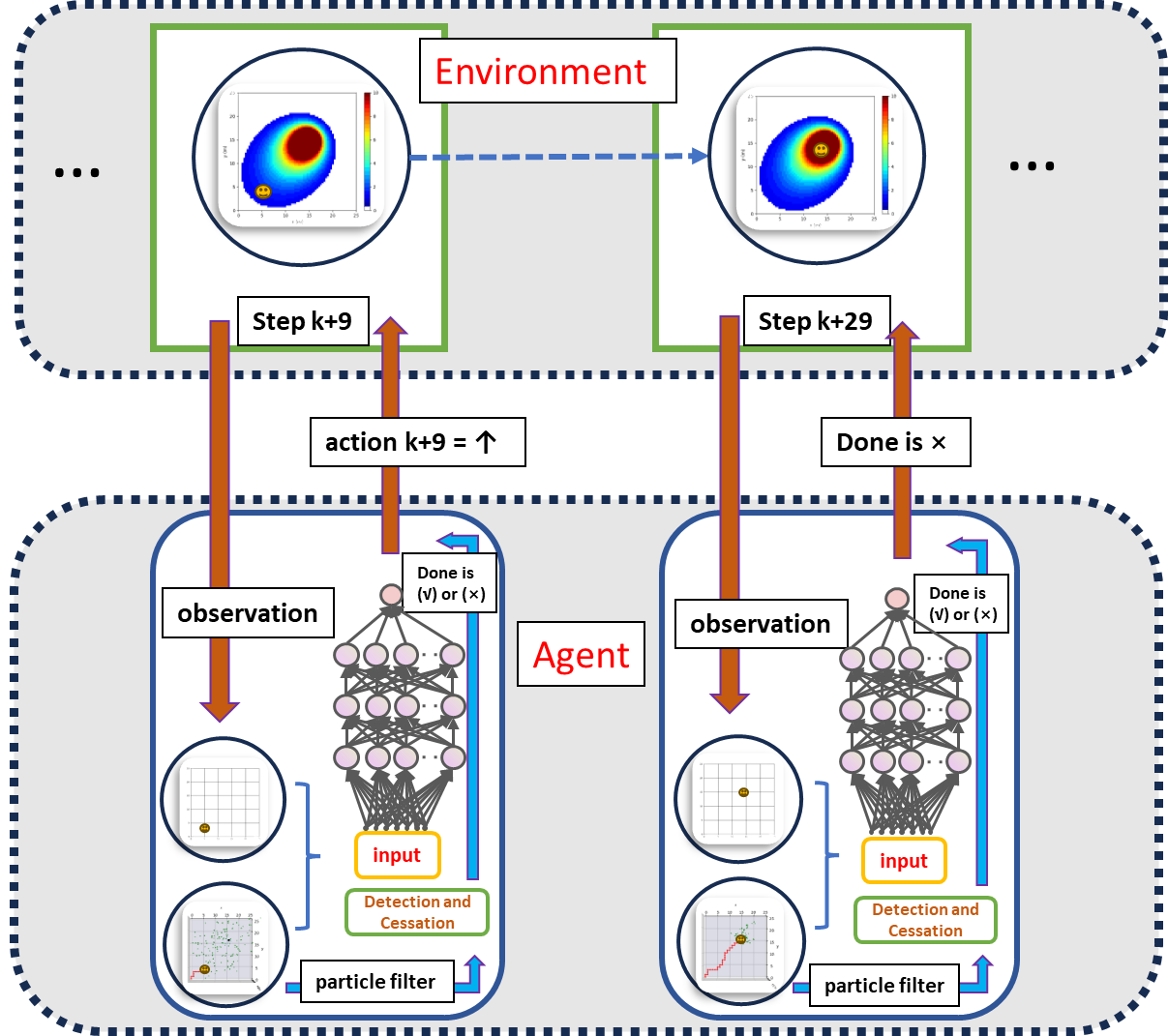}
    \caption{AGDC Structure Diagram for Solving the STE}
    \label{POMDP_STE}
\end{figure}

\textbf{Source Term Estimation} involves identifying the location and characteristics (such as release rate) of hazardous gas emissions in the atmosphere, which is crucial for environmental monitoring and emergency response. However, this task presents several challenges. Firstly, the environment often does not provide clear cessation signals, and gas releases are typically invisible. Secondly, atmospheric turbulence results in sensor data that is sparse, intermittent, and time-varying. Additionally, direct measurements within the hazardous material dispersion area are often dangerous. Furthermore, sensor measurements are highly affected by noise and discontinuity. Consequently, agents must rapidly adjust their strategies in response to environmental changes and accurately estimate the source term despite unclear feedback. These challenges render source term estimation a complex task with significant uncertainty.

Statistical-based methods such as \textit{Infotaxis}, \textit{Entrotaxis} , and \textit{Dual Control for Exploitation and Exploration} (DCEE) provide some solutions to these challenges. Infotaxis in \cite{vergassola2007infotaxis,hutchinson2018information,loisy2022searching} guides the search process by maximizing the rate of information acquisition, gradually reducing uncertainty to approximate the source of the gas. It is suitable for situations with high uncertainty regarding the gas source location but suffers from poor real-time performance and a tendency to get stuck in local optima. Entrotaxis  in \cite{hutchinson2018entrotaxis,zhao2020entrotaxis} minimizes the entropy of future measurements to guide path selection, optimizing information acquisition based on real-time measurement data. It performs robustly in noisy environments but has poor adaptability in dynamic settings or sudden changes, making rapid adjustments difficult. DCEE in \cite{chen2021dual,rhodes2022autonomous} 
 combines strategies of exploitation and exploration by dynamically adjusting the balance between the two, optimizing information acquisition and path selection. Suitable for highly variable environments, this strategy is complex to implement, requiring precise design of the trade-off mechanism between exploration and exploitation. Although these methods are effective in addressing the lack of environmental feedback and clear cessation signals, they share several common limitations: poor real-time performance, poor adaptability in rapidly changing environments, a tendency to get stuck in local optima, and complexity in implementation, particularly for the DCEE.

While Reinforcement Learning offers control capabilities, significantly improving efficiency and success rates, and avoiding local optima with good generalization capabilities, it effectively overcomes the limitations of statistical methods. However, RL struggles to effectively identify and autonomously cease  goals in the absence of clear feedback. \textit{Traditional statistical methods consist of two modules:} one for providing the agent with reward signals based on estimates, where rewards or information gain originate, and the other for determining the optimal action based on information gain. By replacing the control module of traditional statistical methods with RL while retaining the estimation module, a self-feedback mechanism can be introduced to RL. This helps RL algorithms check and cease  goals at appropriate times. Although the rewards for the agent remain sparse in this context, the self-feedback mechanism provides signals from the agent itself rather than the environment, thus better supporting the training process.

Building on this foundation, we propose the concept of Autonomous Goal Detection and Cessation to address the STE problem using RL. AGDC introduces an autonomous goal detection and cessation mechanism within the RL framework, enabling the agent to automatically recognize and cease actions upon task completion. Specifically, the AGDC module employs Bayesian inference to estimate environmental dynamics and dynamically assess task progress. When the standard deviation of the estimated parameter values for environmental dynamics reaches a preset threshold, a cessation signal is automatically triggered. This approach not only addresses the issue of insufficient environmental feedback but also significantly enhances the adaptability and effectiveness of RL in complex and dynamic environments. By integrating the self-feedback mechanism with AGDC, RL algorithms can more efficiently accomplish STE tasks, thereby significantly improving capabilities in environmental monitoring and emergency response. The AGDC structure diagram is shown in Figure \ref{POMDP_STE}.

The main contributions of this paper are as follows: (1) We are the first to introduce, to the best of our knowledge, the AGDC module, which enables agents to autonomously recognize and evaluate goals, significantly increasing the capability of RL algorithms in feedback-limited environments. (2) We demonstrate the successful integration of AGDC with multiple RL algorithms in addressing the STE problem, achieving notable improvements over traditional methods. (3) Extensive experimental results confirm the superiority of AGDC-enhanced RL algorithms in terms of success rate, traveled distance, and search time, particularly in challenging environments with high uncertainty and limited feedback.

\section{Problem Formulation and Peliminaries} \label{sec:Peliminaries}

In a two-dimensional search area denoted as \(\Omega \subseteq \mathbb{R}^2\), where there is an expectation of encountering a hazardous release, a robot equipped with a gas sensor is tasked with traversing the area to calculate the release parameters, also referred to as the source term \(\Theta_s\). This data will serve as the requisite input for a convection-diffusion model \cite{vergassola2007infotaxis}, enabling the generation of hazard forecasts. Within this context, in an environment characterized by average wind speed \(u_s \in \mathbb{R}^+\) in meters per second (m/s), wind direction \(\phi_s\) in radians (rad), and diffusivity \(d_s\) in meters squared per second (m\(^2\)/s), we detect the concentration of a hazardous substance \(x_k \in \mathbb{R}^+\) using a robot or a sensor located at position \(p_k = [x_k, y_k]^T \in \Omega\) in meters (m). This hazardous material stems from a source term located at \(p_s \in \Omega\), and it is released at a rate/strength \(q_s \in \mathbb{R}^+\) in grams per second (g/s) with an average lifespan of \(\tau_s \in \mathbb{R}^+\) in seconds (s).

Consequently, the parameters of the source term can be represented as follows:
\begin{align}
    \Theta_s = [p_s^T, q_s, u_s, \phi_s, d_s, \tau_s]^T
\end{align}

Note: It is assumed in this study that all other parameters, except for the location of the leak source, can be obtained through direct measurement. Therefore, the primary objective is to successfully obtain the source term \(p_s^T\), as it is the most important information about the source term.

\subsection{Gaseous Diffusion Model}
The average gas concentration \cite{zhao2022deep}, denoted as \(m(p|\Theta_s)\), for a given location of the mobile sensor at position \(p_k\) over a time duration of \(\tau_s\), can be computed utilizing the parameters of the source term \(\Theta_s\) as follows:
\begin{align}
m(p_k|\Theta_s) = \frac{q_s}{4\pi d_s \|p_k - p_s\|} \exp\left[\frac{-\|p_k - p_s\|}{\lambda} + \psi\right], \nonumber
\end{align}
where
\begin{align}
\psi &= {-(x_k - x_s) u_s \cos \phi_s}/{2d_s} {-(y_k - y_s) u_s \sin \phi_s}/{2d_s} \nonumber \\
\lambda &= \sqrt{{d_s \tau_s}/{[1 + \left({u_s^2 \tau_s}/{4d_s}\right)}]}. \nonumber
\end{align}

\subsection{Sensor Models}

When mobile robots equipped with sensors conduct gas concentration detection in an environment, the measurement results are influenced by both internal and external factors. Internal factors include the sensitivity and calibration issues of the sensors, as well as the robot's internal dynamics during movement, such as vibrations and posture changes. These factors can significantly impact gas concentration measurements. For example, the sensitivity of metal–oxide gas sensors \cite{li2011odor,neumann2013gas,neumann2015real} can vary with atmospheric contaminants, causing fluctuations in resistance and, consequently, the voltage readings. Furthermore, calibration issues, as discussed in the context of uncalibrated sensors, can lead to inaccuracies in the measured gas concentrations.

External factors encompass environmental variations in gas temperature, humidity, and pressure, all of which can affect measurement results. Wind turbulence, in particular, is a major contributor to noise in sensor readings. This is modeled using Gaussian distributions to describe the uncertainty in gas concentration measurements more accurately. For instance, the noise caused by wind turbulence is represented as \( \nu_{\text{wind}} \sim \mathcal{N}(0, \sigma_{\text{wind}}^2) \), where \( \sigma_{\text{wind}} \) is a parameter that characterizes the instability of the wind conditions.

Specifically, the noise due to internal factors from the sensor is represented as \( v_{\text{inter}} \). Therefore, the value of the gas concentration \( c(p_k|\Theta) \) measured by the robot at position \( p_k \) is $
    c(p_k|\Theta) = m(p_k|\Theta) + v_{\text{inter}}$, where
\( v_{\text{inter}} \sim  \mathcal{N}(u,\sigma) \nonumber \) and  \( v_{\text{inter}} \) follows a normal distribution \( \mathcal{N} \) with mean \( u \) and variance \( \sigma \). This Gaussian noise model is widely used to emulate sensor measurements in simulations, accounting for both wind turbulence and sensor errors.

\begin{figure}[tbhp]
    \centering
    \subfloat[Map of GC model]{
        \label{plume_model}
        \includegraphics[width=0.46\linewidth]{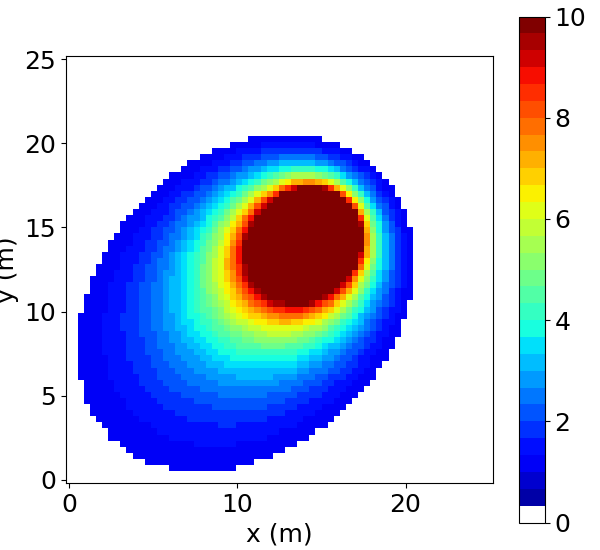}
    }
    \hfill
    \subfloat[Map of GC with sensor noise]{
        \label{sensor_map}
        \includegraphics[width=0.46\linewidth]{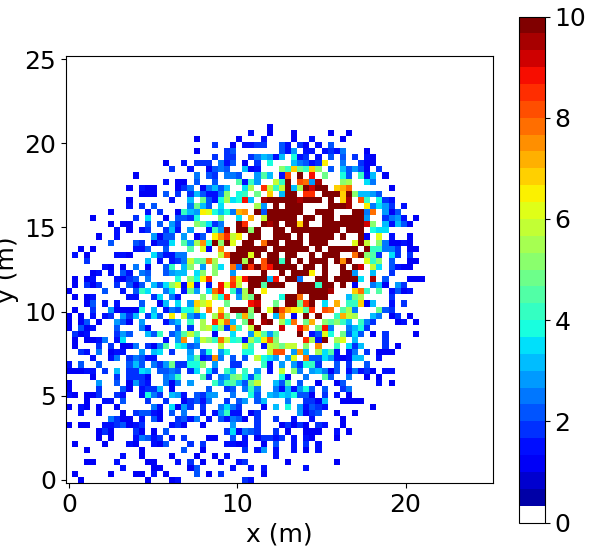}
    }
    \caption{The example of map of the Gas-Diffusion model}
    \label{fig:gas_diffusion_model}
\end{figure}

In a 25-meter by 25-meter area, as an illustrative example, the gaseous diffusion model can provide the gas concentration (GC) at each location. The robot samples the value obtained from the gas concentration model, as shown in Figure (\ref{plume_model}), while also considering the presence of noise in the measured values at each location, depicted in Figure (\ref{sensor_map}). The source term parameters in the gas model are \( q_s = 5 \, \text{g/s}, u_s = 2 \, \text{m/s}, \phi_s = 45^\circ, d_s = 2 \, \text{m}^2/\text{s}, \tau_s = 10 \, \text{s} \).

In the STE problem, the robot can only obtain (or sample) the gas concentration measurements at its current location during the exploration process. Concentrations at other locations are not accessible, highlighting the importance of accurate sensor modeling and noise representation in ensuring reliable gas concentration measurements

\subsection{Partially Observable Markov Decision Process}

In the search area, the robot's task is to estimate the location of a gas leak source with minimal movement steps. The robot lacks prior knowledge of the leak source's location and gathers data through measurements at each step, making decisions based on this information. To address this challenge, we use the Partially Observable Markov Decision Process (POMDP) Reinforcement Learning method. This approach enables the robot to make effective decisions in an environment that is not fully observable, allowing for accurate localization of the gas leak source. A POMDP is a tuple $(S,O,A,T,Pr,R,\gamma)$, where $S$ is a set of states, $O$ is a set of observations, $A$ is a set of actions, $T(s'|s,a) : S \times A \rightarrow S$ is transition probabilities, $Pr(o|s,a) : S \rightarrow O$ is observation probabilities, $R(s,a) : S \times A \rightarrow \mathbb{R}$ is a reward function, and $\gamma$ is a discount factor. The objective in POMDP is to obtain an optimal policy to maximize expected cumulative rewards $\mathbb{E}{[\sum_{k=0}^{\infty} \gamma^k r_k]}$.

\section{Methodology}
\label{sec:Methodology}

\subsection{Bayesian Approximation for Belief Distribution}

The STE challenge is characterized as a Partially Observable Markov Decision Process, which implies that optimal decision-making cannot rely solely on the present observation, as complete state information in environment is not available. To address this issue, we have adopted this approach combining RL and Bayesian inference to effectively estimate  environmental dynamics (belief) distribution and provide more unobservable information to assist decision-making.

Variational inference \cite{jordan1999introduction} and particle filter \cite{gordon1993novel} are common Bayesian inference methods. The former approximates the target distribution by optimizing a differentiable lower bound but typically requires simplifying assumptions that may not suit complex environments. The latter, on the other hand, handles nonlinear and non-Gaussian distributions more effectively and adapts to dynamic conditions without needing such assumptions. Particle filter approximate the true distribution through sufficient sampling and offer flexibility in managing multimodal distributions, tracking potential source locations and release rates, and providing real-time updates. Therefore, particle filter outperform variational inference in STE problems.

The Particle filter is employed for iterative calculations of the environmental dynamics distribution. The belief distribution at time step \(k\) is approximated using a set of \(N\) particles, which are random samples \(\{\Theta_{k}^{i},w_k^{i}\}_{i=1:N}\), where \(\Theta_k^i\) represents the \(i_{th}\) point estimation of source parameters (i.e., the belief), and \(w_k^i\) is the associated weight, with \(\sum_{i=1}^{N} w_k^i = 1\).  The approximated belief distribution $b(\Theta_k)$ is expressed using samples and associated weights as follows:
\begin{align}
b(\Theta_k) = \sum_{i=1}^{N} w_{k}^{i} \delta(\Theta_k - \Theta_k^i),
\end{align}
where \(\delta(\cdot)\) represents the Dirac delta function, which places all its weight at the value \(\Theta_k^i\).

The sampling weights \(\{w_k^{i}\}_{i=1:N}\) are updated through iterative sequential importance sampling \cite{tokdar2010importance}. At each step, a new sample of source parameter estimates \(\{\Theta_{k}^{i}\}_{i=1:N}\) is obtained from the proposal distribution \(q(\Theta_{k}^{i})\). Then, the unnormalized weight \(\bar{w}_{k+1}^{(i)}\) corresponding to the source term vector \(\Theta_{k}\) of the particle filter is updated as follows:
\begin{align}
\bar{w}_{k+1}^{i} &\propto w_{k}^{i} \cdot \frac{Pr(o_{k+1}|\Theta_{k+1}^{i}) ~ T(\Theta_{k+1}^{i}|\Theta_{k}^{i})}{q(\Theta_{k+1}^{i}|\Theta_{k}^{i},{\bf o}_{1:k+1})}.
\end{align}

Obviously, although the dynamic environment and observation results may appear to change due to measurement errors and noise, the parameters constituting the environment (source term) remain fixed, leading to \(\Theta_{k+1} = \Theta_k\) for \(N\) particles. Assuming that the proposal distribution is consistent with the posterior distribution, the formula can be simplified as follows, (see additional materials for details):
\begin{align}
\bar{w}_{k+1}^{i} & = w_{k}^{i} \cdot Pr(o_{k+1}|\Theta_{k+1}^{i}).
\end{align}

Normalization of the sampling weights \(w_k^i\) can be used to approximate the posterior distribution:
\begin{align}
w_{k+1}^{i} & = \frac{w_{k}^{i} \cdot Pr(o_{k+1}|\Theta_{k+1}^{i})}{\sum_{i=1}^{N} w_{k}^{i} \cdot Pr(o_{k+1}|\Theta_{k+1}^{i})} = \frac{\bar{w}_{k+1}^{i}}{\sum_{i=1}^{N}\bar{w}_{k+1}^{i}}.
\end{align}

As the number of learning iterations increases, the weights of most particles tend to zero, leading to the degeneracy problem. To address this issue, resampling is employed, and this method is based on the Markov Chain Monte Carlo move step \cite{Ristic2004BeyondTK}.

We set a resampling threshold \(\epsilon\), and when the number of effective samples falls below this threshold, the resampling process is triggered. The effective samples $N_{eff}$ are calculated by:
\begin{align}
N_{eff} = \frac{1}{\sum_{i=1}^{N} (w_k^i)^2}.
\end{align}

\subsection{Autonomous Goal Detection and Cessation} 
Particle filter not only estimates the distribution of unknown environments but also combines with observational data to provide a more accurate representation of the true state. Additionally, it can generate self-cessation evaluation signals, indicating when computation can be ceased. When the goal is achieved, the particles forming the point estimate converge to a smaller range. By calculating the standard deviation of the particle parameters, the degree of parameter convergence can be observed, facilitating the judgment of goal completion progress.

When the standard deviation (STD) of the belief \(\Theta_k\) by particle filter is lower than the given Cessation Threshold \(\zeta\), the search ceases, and we consider the source term to be successfully estimated, meaning the goal is achieved. As a result, the agent rewards itself rather than relying on the environment. \(STD\) is denoted as:
\begin{align}
STD = \sqrt{diag(Cov(\Theta))},
\end{align}
where \(Cov(\cdot)\) represents the covariance, and \(diag(\cdot)\) denotes the trace of the matrix.

Calculating the STD of the particle filter belief and comparing it with a threshold as a signal to cease  the search process is effective because the standard deviation reflects the concentration level of the belief distribution. A smaller standard deviation indicates that most particles are clustered around a certain estimate, suggesting that the belief state has converged. This convergence implies that the robot's estimation of the source term parameters (such as the leak location) has become more accurate and stable. When the STD is less than the preset threshold \(\zeta\), it indicates that the belief state estimation is sufficiently precise, making further algorithm execution yield diminishing returns, thus allowing the search process to cease, saving computational resources and time. Additionally, using the standard deviation as a cessation condition is relatively simple and intuitive, making it easier to implement and understand, facilitating easier decision-making and control in practical applications.
\begin{algorithm}[htbp]
\caption{Reinforcement Learning with AGDC}
\begin{algorithmic}[1]
    \STATE Initialize environment $\mathcal{E}$, policy $\pi$ and value function $V$, particle filter with $N$ particles
    \FOR{episode = 1 to $M$}
        \STATE Initialize observation $o_0$ from $\mathcal{E}$, particles $\{\Theta\}_{i=1}^N$, state estimate $\hat{s}_0$  from particles $\{\Theta_k^i\}_{i=1}^N$ and observation $o_0$
        \FOR{ time step = 1 to $k$}
            \STATE Sample $a_k \sim \pi_\theta(s_k)$, $o_k \sim \mathcal{E}$ 
            
            \STATE Update particles $\{\Theta_k^i\}_{i=1}^N$ using particle filter
                \FOR{each particle $i$}
                    \STATE Sample new particle $\Theta_{k+1}^i \sim T(s_{k+1} | s_{k}, a_k)$
                    \STATE Compute weight $w_{k+1}^i = w_{k}^i\cdot Pr(s_{k+1} | \Theta_{k+1}^i)$

                    \IF{$N_{eff} < \epsilon $}
                        \STATE Normalize weights $w_t^{(i)} = \frac{w_t^{(i)}}{\sum_{j=1}^{N_p} w_t^{(j)}}$
                        \STATE Resample  $\{\Theta_{k+1}^i\}_{i=1}^N$ based on $\{w_{k+1}^i\}_{i=1}^N$
                    \ENDIF
        
                \ENDFOR

            \STATE $r_k \leftarrow$ 0,  $done \leftarrow \text{False}$
                
            \IF{$STD < \zeta$}
                \STATE $r_k \leftarrow$ value$>0$, $done \leftarrow \text{True}$
            \ENDIF
            
            \STATE Estimate state  $\hat{s}_{k+1}$ from $\{\Theta_k^i\}_{i=1}^N$ and  $o_k$
            \STATE Update policy $\pi$  and $V$ using $\hat{s}_k$, $a_k$, $r_k$, and $\hat{s}_{k+1}$
            \STATE Set $o_k \leftarrow o_{k+1}$, $\hat{s}_k \leftarrow \hat{s}_{k+1}$
        \ENDFOR
    \ENDFOR
\end{algorithmic}
\label{AGDC}
\end{algorithm}

\begin{figure*}[!ht]
    \centering
    \begin{subfigure}[b]{0.32\linewidth}
        \centering        \includegraphics[width=\linewidth,height=0.8\linewidth]{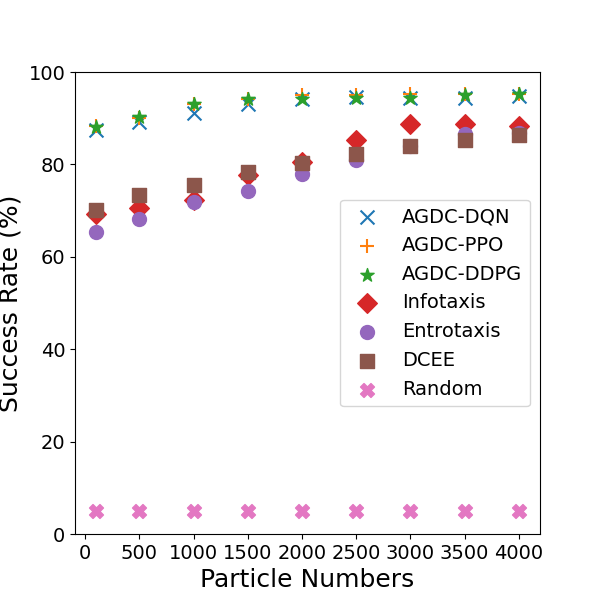}
        \caption{Success Rates Across PN}
    \end{subfigure}
    \hfill
    \begin{subfigure}[b]{0.32\linewidth}
        \centering        \includegraphics[width=\linewidth,height=0.8\linewidth]{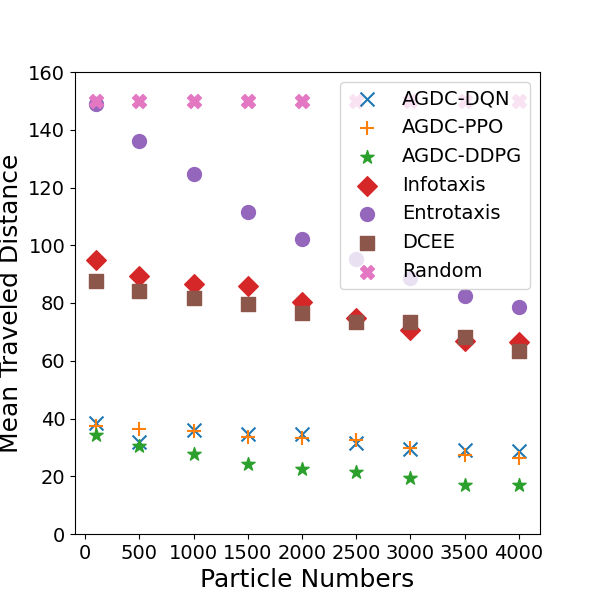}
        \caption{Mean Traveled Distance Across PN}
    \end{subfigure}
    \hfill
    \begin{subfigure}[b]{0.33\linewidth}
        \centering        \includegraphics[width=\linewidth,height=0.8\linewidth]{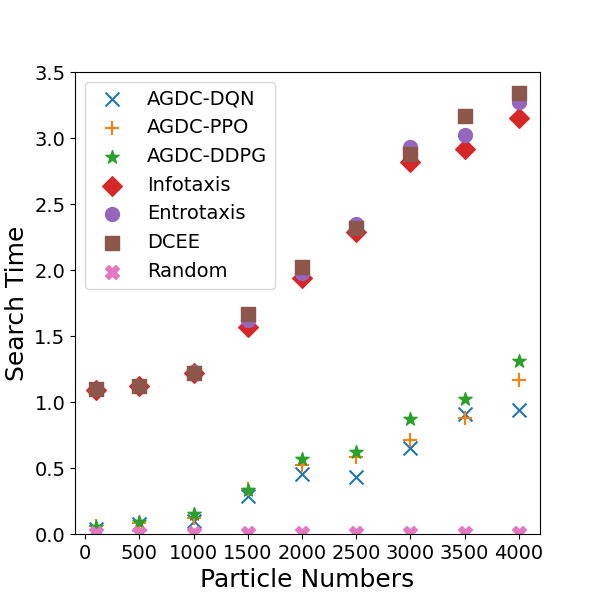}
        \caption{Search Time Across PN}
    \end{subfigure}
    
    \vskip\baselineskip
    
    \begin{subfigure}[b]{0.33\linewidth}
        \centering        \includegraphics[width=\linewidth,height=0.8\linewidth]{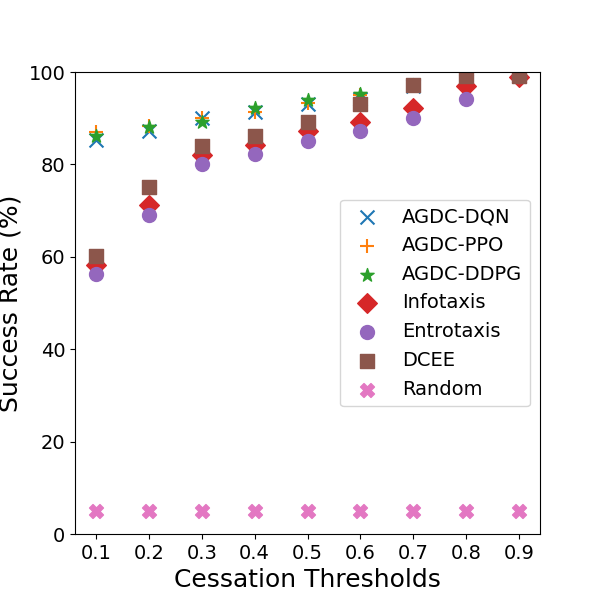}
        \caption{Success Rates Across CT}
    \end{subfigure}
    \hfill
    \begin{subfigure}[b]{0.33\linewidth}
        \centering        \includegraphics[width=\linewidth,height=0.8\linewidth]{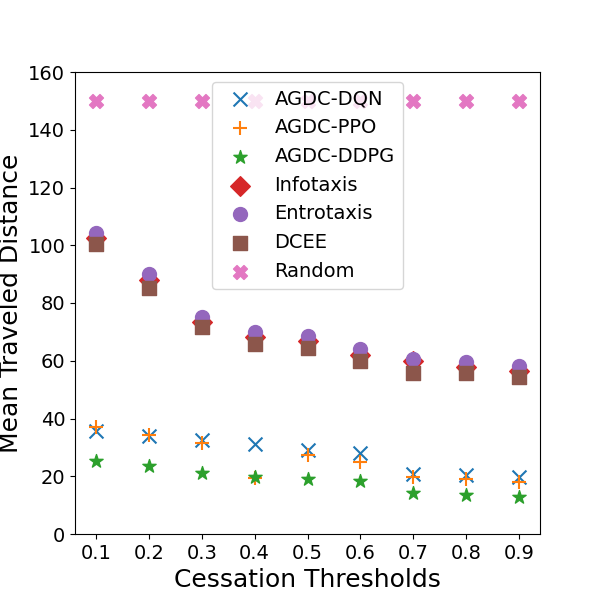}
        \caption{Mean Traveled Distance Across CT}
    \end{subfigure}
    \hfill
    \begin{subfigure}[b]{0.32\linewidth}
        \centering        \includegraphics[width=\linewidth,height=0.8\linewidth]{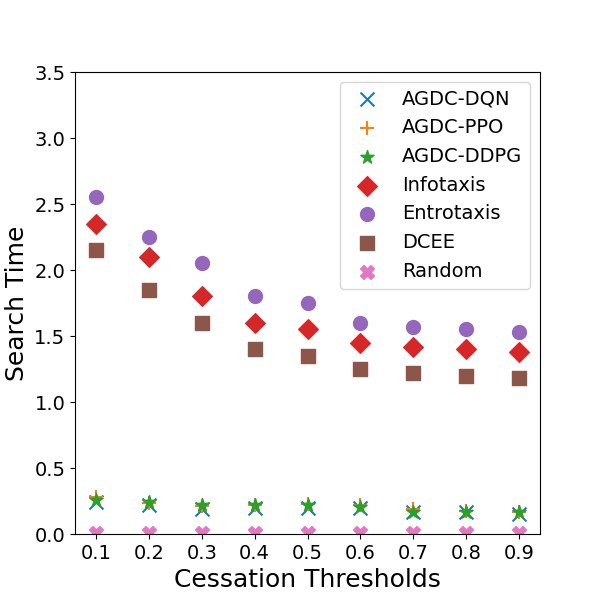}
        \caption{Search Time Across CT}
    \end{subfigure}
    
    \caption{Comparative Analysis of Particle Numbers (PN) and Cessation Thresholds(CT)}
    \label{fig:comparison}
\end{figure*}

For example, suppose the covariance matrix of the particle filter belief state is $
Cov(\Theta) = \begin{pmatrix}
\sigma_x^2 & \sigma_{xy} \\
\sigma_{xy} & \sigma_y^2
\end{pmatrix}$. Using \textit{diag($\cdot)$}, the standard deviations for the \(x\) and \(y\) coordinates can be calculated as $STD_{x/y}  = \sqrt{\sigma_{x/y}^2} $. This approach allows for an intuitive assessment of the uncertainty for each parameter. In contrast, directly calculating the standard deviation of the entire covariance matrix would require complex matrix operations, making the results less intuitive to understand. Therefore, using \textit{diag($\cdot$)} for standard deviation calculation is a more effective and intuitive method. The details of the training algorithm
are shown in Algorithm \ref{AGDC}.

\section{Experiment} \label{sec:Experiment}

In this paper, we utilized the STE Environment (STEenv) as the experimental platform to investigate the application of AGDC in RL. The STE Environment, described in Section 2 or \cite{shi2024reinforcement}, comprises a Gaussian diffusion model and a sensor model. Instead of providing rewards to the Agent, it transmits the current position and the test concentration at that position as observations. Any RL algorithm can be integrated with the AGDC module to address the challenges posed by the STE.

\subsection{Baseline Algorithms and Evaluation Metrics}

There are three evaluation metrics in this paper: \textit{Success Rate} (SR) to measure the number of successful source term estimations in the trajectories,  \textit{Mean Traveled Distance} (MTD) to represent the average distance traveled before a successful estimation occurs, and \textit{Search Time} (ST) to measure the duration from the initiation of a search to the successful estimation of the target information. In this context, achieving a shorter traveled distance while successfully estimating the source's location more frequently, along with a shorter search time indicating a quicker and more effective search process, signifies a more efficient approach.

We integrate the AGDC module into the DQN, PPO, and DDPG algorithms to address the STE problem. Additionally, we use four baseline approaches: Infotaxis, Entrotaxis, Dual Control for Exploitation and Exploration (DCEE), and a method that randomly selects actions for control. The first three are statistical methods, while the last one is a non-statistical method based on random action selection.

\noindent \textbf{Infotaxis}  \cite{vergassola2007infotaxis} aims to minimize the predicted posterior variance of the source location. It treats the search process as an information-gathering problem, where the objective is to reduce uncertainty about the source location.

\noindent  \textbf{Entrotaxis} \cite{hutchinson2018entrotaxis} enforces the maximum entropy sampling principle, which directs the agent to move towards positions of maximum uncertainty to gather more informative data about the environment and the source location.

\noindent \textbf{DCEE} \cite{chen2021dual} integrates both exploitation and exploration by incorporating the uncertainty into the control decisions, thus allowing the robot to not only move towards the estimated target but also probe the environment to reduce uncertainty.

\noindent \textbf{Random} involves selecting actions randomly, without any strategic guidance or optimization. It serves as a non-statistical baseline to compare against more sophisticated methods, highlighting the benefits of informed and strategic action selection.

\begin{table}[htbp]
    \centering

    \begin{tabular}{|c|c|}
        \hline
        \textbf{Source Parameter} & \textbf{Distribution} \\ \hline
        Source Location \( x_s\) & Uniform \( \mathcal{U}(10, 25) \) \\ \hline
        Source Location \( y_s \) & Uniform \( \mathcal{U}(10, 25) \) \\ \hline
        Release Strength \( q_s \) & Uniform \( \mathcal{U}(100, 500)\) \\ \hline
        Wind Speed \( u_s \) & Uniform \( \mathcal{U}(1, 4) \) \\ \hline
        Wind Direction \( \phi_s \) & Uniform \( \mathcal{U}(0, 360) \)\\ \hline
        Diffusivity \( d \) & Uniform \( \mathcal{U}(1, 8) \)\\ \hline
        Sensor Noise \( \alpha \) & Fixed at 0.3 \\ \hline
        Environmental Noise \( \beta \) & Fixed at 0.2 \\ \hline
        Effective Samples $N_{eff}$ \( \beta \) & Fixed at 0.5 \\ \hline
    \end{tabular}
    \caption{Parameter Distributions for the Training Scenarios}
    \label{tab:training_parameters}
\end{table}
\subsection{Scenario Parameterization and Evaluation}

The parameters for the training scenarios, set within a \(30 \times 30\) area, are constructed by randomly initializing the source and environmental properties at the beginning of each training episode, including parameters such as the gas source location, wind speed, and wind direction, all of which are sampled from the probability distributions presented in Table \ref{tab:training_parameters}. The agent starts its search from a random location within the \((0, 5) \times (0, 5)\) area, with a speed of 1 meter per step, ensuring that it encounters a diverse set of scenarios during training, thereby promoting robust learning. The parameters for the testing scenarios consist of 1,000 randomly generated conditions that are not used during training. Although these scenarios are generated from the same parameter ranges as the training data, they present different specific conditions to ensure that the model is evaluated on unseen data.

\subsection{Particle Numbers and Cessation Thresholds}
\begin{figure}[htbp]
    \centering
    \includegraphics[width=0.95\linewidth]{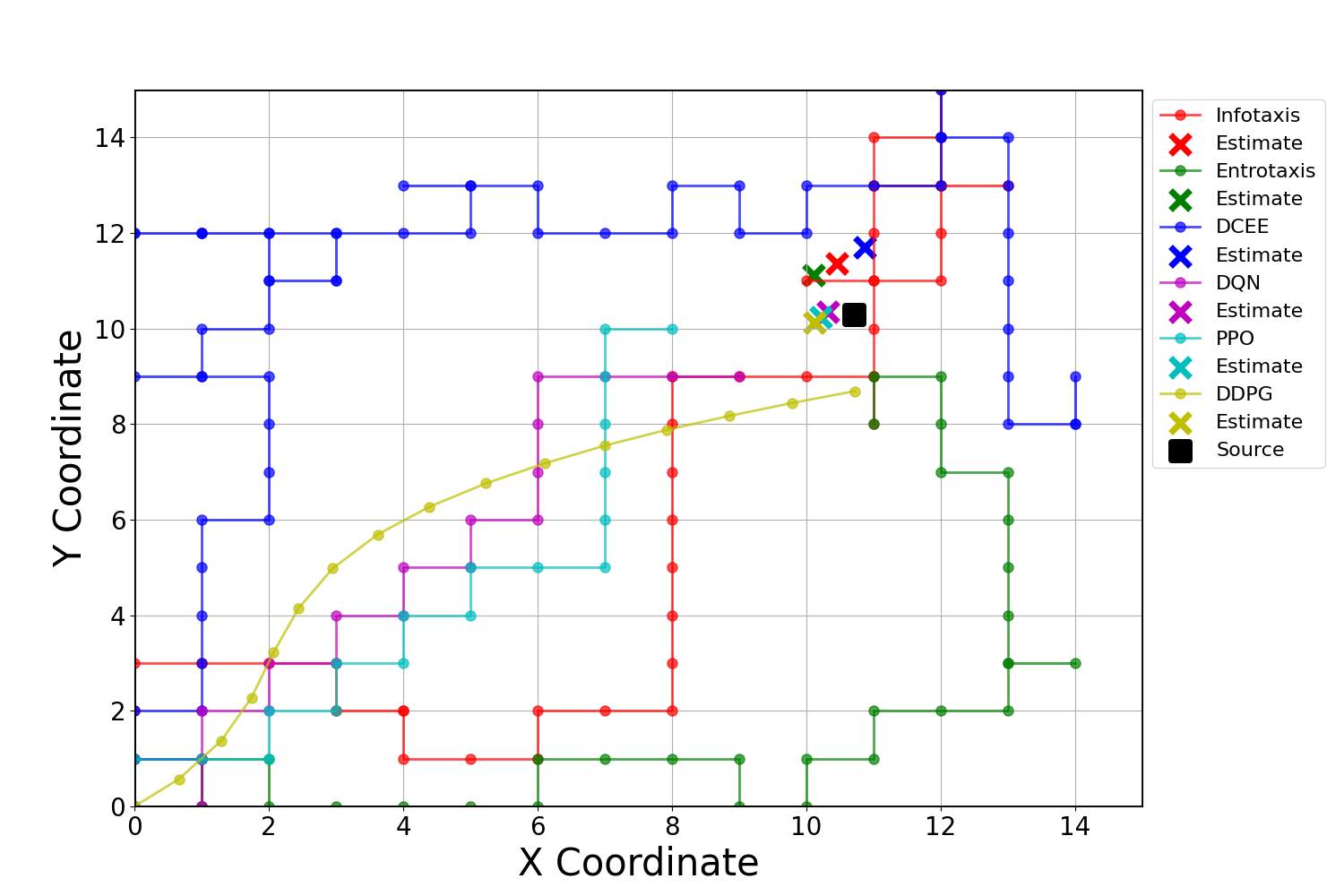}
    \caption{Trajectories of Various Methods}
    \label{mutli-step_7mothed}
\end{figure}

We will discuss the effectiveness of integrating the AGDC module with RL by comparing three AGDC-enhanced RL algorithms against four baseline methods. Before delving into these comparisons, it is important to introduce two critical hyperparameters: Particle Numbers (PN) and Cessation Thresholds (CT) $\zeta$. The Particle Numbers are pivotal for AGDC as they determine the number of samples used to approximate the point distribution, which is essential for accurately estimating the belief necessary for goal detection. The Cessation Thresholds are directly linked to the cessation aspect of AGDC, as they define the criteria for deciding when to stop the task, ensuring that the process concludes at the most appropriate moment. Two groups of experiments were conducted to investigate the impact of two key hyperparameters on the performance of different algorithms. The first group analyzed the variations in success rate, mean traveled distance, and search time as the number of particles increased from 100 to 4000. The second group examined algorithm performance under different cessation thresholds, ranging from 0.1 to 0.9. The results are presented in Figure \ref{fig:comparison}. The trajectories of six methods (excluding the random method) are shown in Figure \ref{mutli-step_7mothed}.

\noindent \textbf{Analysis Across Various Particle Numbers}: The success rates of AGDC-enhanced RL algorithms, such as AGDC-DQN, AGDC-PPO, and AGDC-DDPG, improve steadily with increasing particle numbers, reaching 95.27\% for AGDC-PPO at 4000 particles. These algorithms outperform baseline methods like Infotaxis, Entrotaxis, and DCEE, which also benefit from more particles but remain less effective. The Random method remains largely ineffective, with success rates below 5\%, regardless of particle count. Notably, DDPG's continuous action space allows for more flexible  movement compared to the discrete steps of other methods. AGDC-enhanced algorithms also show a consistent reduction in mean traveled distance as particle numbers increase. For instance, AGDC-DDPG reduces its mean distance from 34.33 at 100 particles to 16.83 at 4000, demonstrating greater efficiency. In contrast, baseline methods like Entrotaxis exhibit significantly higher distances, highlighting their inefficiency. The Random method, unaffected by particle count, continues to exhibit very high traveled distances. Regarding search time, AGDC-enhanced algorithms consistently achieve quicker convergence compared to baseline methods. AGDC-DQN, for example, maintains a search time of 0.94 units at 4000 particles. In contrast, methods like Infotaxis take significantly longer, reaching 3.15 units, indicating slower and less effective strategies. The Random method, despite its speed, fails to produce meaningful results, underscoring its overall inefficiency.

\noindent \textbf{Analysis Across Various Cessation Thresholds}: The success rates of AGDC-enhanced algorithms, including AGDC-DQN, AGDC-PPO, and AGDC-DDPG, improve with higher cessation thresholds, with AGDC-PPO achieving 99.13\% at a threshold of 0.9. This is due to the algorithms accumulating more confidence before stopping, leading to greater accuracy. Although Infotaxis and Entrotaxis also improve with higher thresholds, they require significantly more time and distance to reach similar success levels. As cessation thresholds rise, AGDC-enhanced algorithms also show reduced mean traveled distances, reflecting more efficient search cessation. AGDC-DDPG, for instance, reduces its distance to 12.8 units at a threshold of 0.9, while Infotaxis, even at its best, remains at 56.4 units, much higher than AGDC-DDPG. Search times for AGDC-enhanced algorithms decrease with higher thresholds, with AGDC-PPO reducing its time to 0.17 units at a threshold of 0.9. This indicates that these algorithms can efficiently conclude the search once confident in their estimation. \textbf{However, excessively high thresholds may artificially inflate success rates and reduce traveled distances, creating a misleading appearance of improved performance by lowering the task’s rigor. Therefore, we advocate for setting an appropriate threshold that balances accuracy and task difficulty.}

\subsection{Further Research Experiments}
\begin{figure}[htbp]
    \centering
    \includegraphics[width=0.95\linewidth]{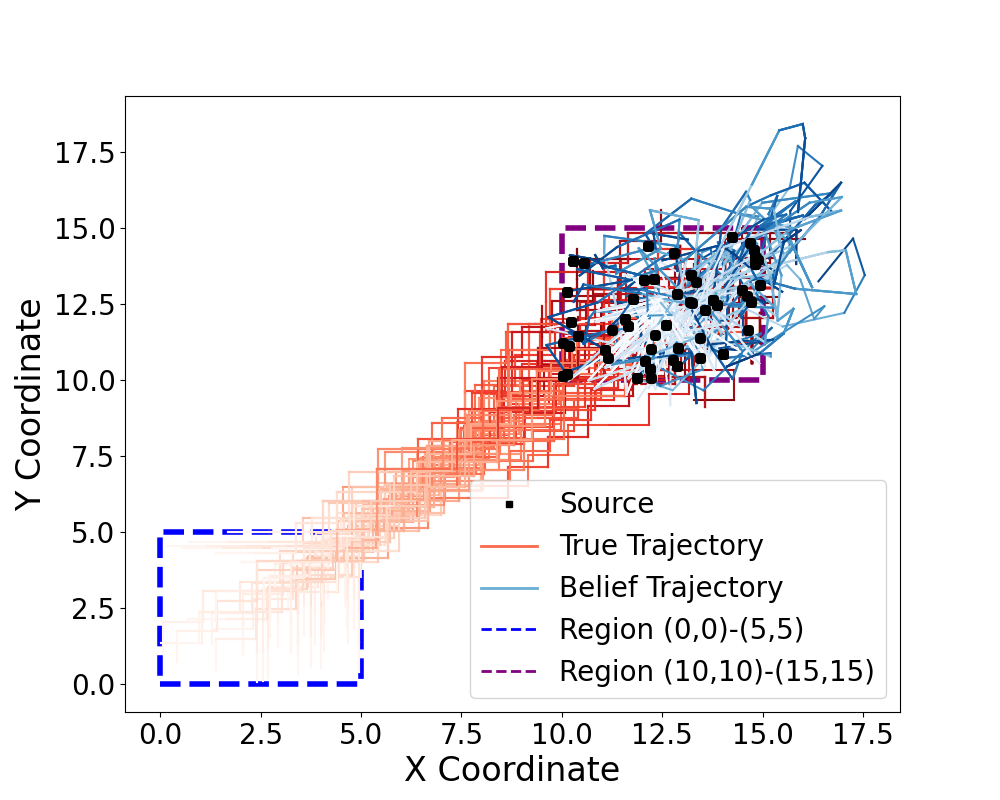}
    \caption{Multiple Trajectories from True Trajectory and Estimated (Belief) Trajectory with Color Gradient}
    \label{mutli-step-no-color}
\end{figure}

\begin{figure}[htbp]
    \centering
    \includegraphics[width=0.95\linewidth]{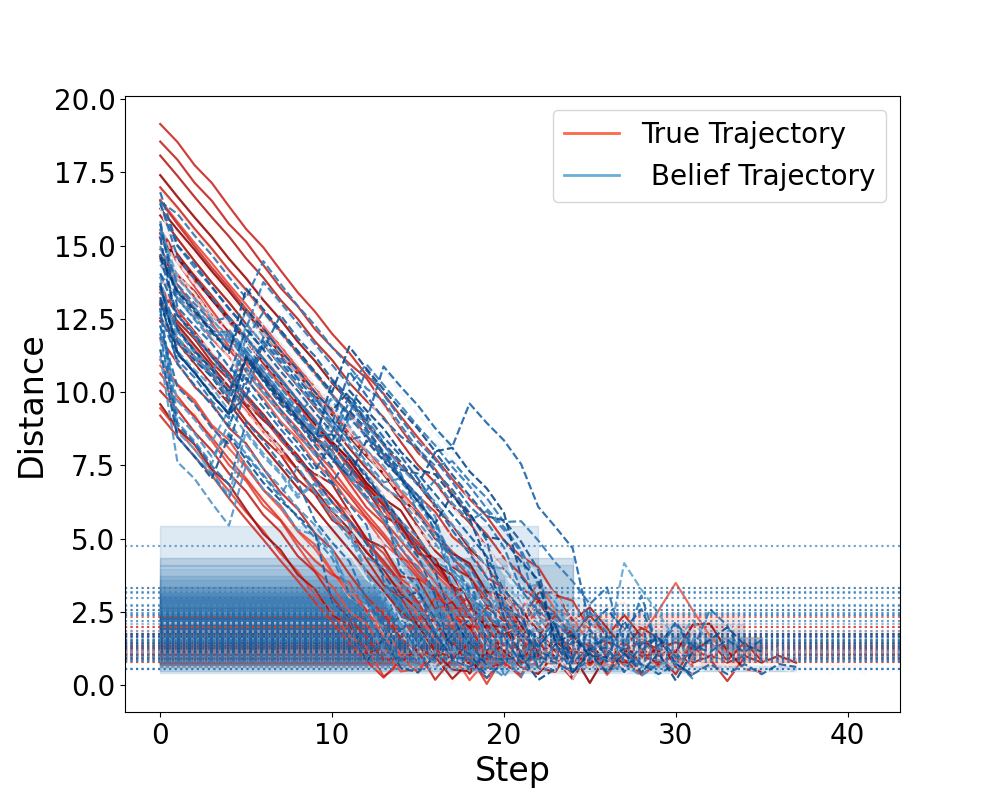}
    \caption{Distance from current position to goal and estimated belief at each step}
    \label{dis_from_cur_goal}
\end{figure}

An additional experiment was conducted to explore the AGDC module through a visualization approach. This experiment uses only the AGDC-based QDN method, with the environment set within a 20×20 area, and the source randomly located within the x, y from $\mathcal{U}(15, 20)$. The experimental results are shown in Figure \ref{mutli-step-no-color} and \ref{dis_from_cur_goal}. In Figure \ref{mutli-step-no-color}, 100 trajectories are displayed, with the red lines representing the agent's actual paths. The color intensity correlates with the time step count, with darker lines indicating more steps. The blue lines show the changes in the estimated source position. As the agent gets closer to the true source location (the goal), the estimated position (derived from Belief) also becomes more accurate, converging towards the goal.

Similarly, in Figure \ref{dis_from_cur_goal}, the red line represents the distance between the agent and the true goal at each step, while the blue line shows the distance between the agent and the current estimated position. The red line in Figure \ref{dis_from_cur_goal} is relatively smooth, indicating that the agent consistently moves towards the true goal, even without knowing its exact location. This can explain why AGDC-based RL methods are efficient, as they avoid taking redundant steps. During this process, the estimated position (Belief) fluctuates with the incoming information but eventually converges near the true goal. We observed that towards the end, the blue line closely aligns with the red line, indicating high estimation accuracy. By setting different CT $\zeta$, represented by the horizontal blue lines, we can determine varying success rates based on the number of red line points below the threshold. This explains why a larger threshold increases the success rate; however, when set too high, the prediction becomes distorted. Combining insights from both figures, we conclude that the agent can effectively achieve Autonomous Goal Detection and Cessation in a gradual manner. Even though the red line may not be available in real-time, the trend of the blue line can still be used to roughly estimate the current progress of recognition, which serves as a form of explainability for ADGC.

\section{Conclusion} \label{sec:Conclusion}

In this study, we introduced the concept of Autonomous Goal Detection and Cessation within the Reinforcement Learning framework to address the challenges of STE. By integrating Bayesian inference with Reinforcement Learning, AGDC enhances the agent's ability to autonomously detect and cease actions upon achieving the goal, significantly improving adaptability and efficiency in complex and dynamic environments. Our experiments demonstrate that AGDC-based methods consistently outperform traditional statistical approaches in terms of success rate, traveled distance, and search time across varying scenarios. This indicates that AGDC provides a robust mechanism for agents to effectively navigate uncertain environments, making it a valuable tool for applications in environmental monitoring and emergency response. Future work could explore the scalability of AGDC to more complex multi-agent systems and real-world deployment scenarios.

\bibliography{aaai25}

\begin{thebibliography}{55}
\providecommand{\natexlab}[1]{#1}

\bibitem[{Anderson and Moore(2005)}]{anderson2005optimal}
Anderson, B.~D.; and Moore, J.~B. 2005.
\newblock \emph{Optimal filtering}.
\newblock Courier Corporation.

\bibitem[{Bai et~al.(2023)Bai, Zhang, Tao, Wu, Wang, and Xu}]{Bai_Zhang_Tao_Wu_Wang_Xu_2023}
Bai, F.; Zhang, H.; Tao, T.; Wu, Z.; Wang, Y.; and Xu, B. 2023.
\newblock PiCor: Multi-Task Deep Reinforcement Learning with Policy Correction.
\newblock \emph{Proceedings of the AAAI Conference on Artificial Intelligence}, 37(6): 6728--6736.

\bibitem[{Bai et~al.(2024)Bai, Zhao, Zhang, Cui, Wen, Yang, Xu, and Han}]{bai2024efficient}
Bai, F.; Zhao, R.; Zhang, H.; Cui, S.; Wen, Y.; Yang, Y.; Xu, B.; and Han, L. 2024.
\newblock Efficient Preference-based Reinforcement Learning via Aligned Experience Estimation.
\newblock \emph{arXiv preprint arXiv:2405.18688}.

\bibitem[{Bollacker et~al.(2008)Bollacker, Evans, Paritosh, Sturge, and Taylor}]{bollacker2008freebase}
Bollacker, K.; Evans, C.; Paritosh, P.; Sturge, T.; and Taylor, J. 2008.
\newblock Freebase: a collaboratively created graph database for structuring human knowledge.
\newblock In \emph{Proceedings of the 2008 ACM SIGMOD international conference on Management of data}, 1247--1250.

\bibitem[{Chen, Rhodes, and Liu(2021)}]{chen2021dual}
Chen, W.-H.; Rhodes, C.; and Liu, C. 2021.
\newblock Dual control for exploitation and exploration (DCEE) in autonomous search.
\newblock \emph{Automatica}, 133: 109851.

\bibitem[{Chui, Chen et~al.(2017)}]{chui2017kalman}
Chui, C.~K.; Chen, G.; et~al. 2017.
\newblock \emph{Kalman filtering}.
\newblock Springer.

\bibitem[{Degrave et~al.(2022)Degrave, Felici, Buchli, Neunert, Tracey, Carpanese, Ewalds, Hafner, Abdolmaleki, de~Las~Casas et~al.}]{degrave2022magnetic}
Degrave, J.; Felici, F.; Buchli, J.; Neunert, M.; Tracey, B.; Carpanese, F.; Ewalds, T.; Hafner, R.; Abdolmaleki, A.; de~Las~Casas, D.; et~al. 2022.
\newblock Magnetic control of tokamak plasmas through deep reinforcement learning.
\newblock \emph{Nature}, 602(7897): 414--419.

\bibitem[{Doucet et~al.(2001)Doucet, De~Freitas, Gordon et~al.}]{doucet2001sequential}
Doucet, A.; De~Freitas, N.; Gordon, N.~J.; et~al. 2001.
\newblock \emph{Sequential Monte Carlo methods in practice}, volume~1.
\newblock Springer.

\bibitem[{Feng et~al.(2023)Feng, Sun, Yan, Zhu, Zou, Shen, and Liu}]{feng2023dense}
Feng, S.; Sun, H.; Yan, X.; Zhu, H.; Zou, Z.; Shen, S.; and Liu, H.~X. 2023.
\newblock Dense reinforcement learning for safety validation of autonomous vehicles.
\newblock \emph{Nature}, 615(7953): 620--627.

\bibitem[{Geman and Geman(1984)}]{geman1984stochastic}
Geman, S.; and Geman, D. 1984.
\newblock Stochastic relaxation, Gibbs distributions, and the Bayesian restoration of images.
\newblock \emph{IEEE Transactions on pattern analysis and machine intelligence}, (6): 721--741.

\bibitem[{Gordon, Salmond, and Smith(1993)}]{gordon1993novel}
Gordon, N.~J.; Salmond, D.~J.; and Smith, A.~F. 1993.
\newblock Novel approach to nonlinear/non-Gaussian Bayesian state estimation.
\newblock In \emph{IEE proceedings F (radar and signal processing)}, volume 140, 107--113. IET.

\bibitem[{Graves and Graves(2012)}]{graves2012long}
Graves, A.; and Graves, A. 2012.
\newblock Long short-term memory.
\newblock \emph{Supervised sequence labelling with recurrent neural networks}, 37--45.

\bibitem[{Hutchinson, Liu, and Chen(2018)}]{hutchinson2018information}
Hutchinson, M.; Liu, C.; and Chen, W.-H. 2018.
\newblock Information-based search for an atmospheric release using a mobile robot: Algorithm and experiments.
\newblock \emph{IEEE Transactions on Control Systems Technology}, 27(6): 2388--2402.

\bibitem[{Hutchinson, Oh, and Chen(2018)}]{hutchinson2018entrotaxis}
Hutchinson, M.; Oh, H.; and Chen, W.-H. 2018.
\newblock Entrotaxis as a strategy for autonomous search and source reconstruction in turbulent conditions.
\newblock \emph{Information Fusion}, 42: 179--189.

\bibitem[{Jordan et~al.(1999)Jordan, Ghahramani, Jaakkola, and Saul}]{jordan1999introduction}
Jordan, M.~I.; Ghahramani, Z.; Jaakkola, T.~S.; and Saul, L.~K. 1999.
\newblock An introduction to variational methods for graphical models.
\newblock \emph{Machine learning}, 37: 183--233.

\bibitem[{Julier and Uhlmann(1997)}]{julier1997new}
Julier, S.~J.; and Uhlmann, J.~K. 1997.
\newblock New extension of the Kalman filter to nonlinear systems.
\newblock In \emph{Signal processing, sensor fusion, and target recognition VI}, volume 3068, 182--193. Spie.

\bibitem[{Kalman(1960)}]{kalman1960new}
Kalman, R.~E. 1960.
\newblock A new approach to linear filtering and prediction problems.

\bibitem[{Kiran et~al.(2021)Kiran, Sobh, Talpaert, Mannion, Al~Sallab, Yogamani, and P{\'e}rez}]{kiran2021deep}
Kiran, B.~R.; Sobh, I.; Talpaert, V.; Mannion, P.; Al~Sallab, A.~A.; Yogamani, S.; and P{\'e}rez, P. 2021.
\newblock Deep reinforcement learning for autonomous driving: A survey.
\newblock \emph{IEEE Transactions on Intelligent Transportation Systems}, 23(6): 4909--4926.

\bibitem[{Kulkarni et~al.(2016)Kulkarni, Narasimhan, Saeedi, and Tenenbaum}]{kulkarni2016hierarchical}
Kulkarni, T.~D.; Narasimhan, K.; Saeedi, A.; and Tenenbaum, J. 2016.
\newblock Hierarchical deep reinforcement learning: Integrating temporal abstraction and intrinsic motivation.
\newblock \emph{Advances in neural information processing systems}, 29.

\bibitem[{Lassila, Hendler, and Berners-Lee(2001)}]{lassila2001semantic}
Lassila, O.; Hendler, J.; and Berners-Lee, T. 2001.
\newblock The semantic web.
\newblock \emph{Scientific American}, 284(5): 34--43.

\bibitem[{Levine et~al.(2016)Levine, Finn, Darrell, and Abbeel}]{levine2016end}
Levine, S.; Finn, C.; Darrell, T.; and Abbeel, P. 2016.
\newblock End-to-end training of deep visuomotor policies.
\newblock \emph{Journal of Machine Learning Research}, 17(39): 1--40.

\bibitem[{Li et~al.(2011)Li, Meng, Wang, and Zeng}]{li2011odor}
Li, J.-G.; Meng, Q.-H.; Wang, Y.; and Zeng, M. 2011.
\newblock Odor source localization using a mobile robot in outdoor airflow environments with a particle filter algorithm.
\newblock \emph{Autonomous Robots}, 30: 281--292.

\bibitem[{Li et~al.(2015)Li, Li, Ji, and Dai}]{li2015kalman}
Li, Q.; Li, R.; Ji, K.; and Dai, W. 2015.
\newblock Kalman filter and its application.
\newblock In \emph{2015 8th international conference on intelligent networks and intelligent systems (ICINIS)}, 74--77. IEEE.

\bibitem[{Li, Wang, and Shi(2023)}]{li2023passivity}
Li, R.; Wang, J.-L.; and Shi, Y.-W. 2023.
\newblock Passivity-based formation control for second-order multi-agent systems with linear or nonlinear coupling.
\newblock \emph{International Journal of Control}, 96(5): 1190--1201.

\bibitem[{Li et~al.(2023)Li, Zhang, Sun, Du, Wen, Wang, and Pan}]{li2023cooperative}
Li, Y.; Zhang, S.; Sun, J.; Du, Y.; Wen, Y.; Wang, X.; and Pan, W. 2023.
\newblock Cooperative open-ended learning framework for zero-shot coordination.
\newblock In \emph{International Conference on Machine Learning}, 20470--20484. PMLR.

\bibitem[{Li et~al.(2024)Li, Zhang, Sun, Zhang, Du, Wen, Wang, and Pan}]{li2024tackling}
Li, Y.; Zhang, S.; Sun, J.; Zhang, W.; Du, Y.; Wen, Y.; Wang, X.; and Pan, W. 2024.
\newblock Tackling cooperative incompatibility for zero-shot human-ai coordination.
\newblock \emph{Journal of Artificial Intelligence Research}, 80: 1139--1185.

\bibitem[{Liu et~al.(2022)Liu, Bai, Du, and Yang}]{NEURIPS2022_8be9c134}
Liu, R.; Bai, F.; Du, Y.; and Yang, Y. 2022.
\newblock Meta-Reward-Net: Implicitly Differentiable Reward Learning for Preference-based Reinforcement Learning.
\newblock In Koyejo, S.; Mohamed, S.; Agarwal, A.; Belgrave, D.; Cho, K.; and Oh, A., eds., \emph{Advances in Neural Information Processing Systems}, volume~35, 22270--22284. Curran Associates, Inc.

\bibitem[{Loisy and Eloy(2022)}]{loisy2022searching}
Loisy, A.; and Eloy, C. 2022.
\newblock Searching for a source without gradients: how good is infotaxis and how to beat it.
\newblock \emph{Proceedings of the Royal Society A}, 478(2262): 20220118.

\bibitem[{Metropolis et~al.(1953)Metropolis, Rosenbluth, Rosenbluth, Teller, and Teller}]{metropolis1953equation}
Metropolis, N.; Rosenbluth, A.~W.; Rosenbluth, M.~N.; Teller, A.~H.; and Teller, E. 1953.
\newblock Equation of state calculations by fast computing machines.
\newblock \emph{The journal of chemical physics}, 21(6): 1087--1092.

\bibitem[{Meyes et~al.(2017)Meyes, Tercan, Roggendorf, Thiele, B{\"u}scher, Obdenbusch, Brecher, Jeschke, and Meisen}]{meyes2017motion}
Meyes, R.; Tercan, H.; Roggendorf, S.; Thiele, T.; B{\"u}scher, C.; Obdenbusch, M.; Brecher, C.; Jeschke, S.; and Meisen, T. 2017.
\newblock Motion planning for industrial robots using reinforcement learning.
\newblock \emph{Procedia CIRP}, 63: 107--112.

\bibitem[{Mnih et~al.(2015)Mnih, Kavukcuoglu, Silver, Rusu, Veness, Bellemare, Graves, Riedmiller, Fidjeland, Ostrovski et~al.}]{mnih2015human}
Mnih, V.; Kavukcuoglu, K.; Silver, D.; Rusu, A.~A.; Veness, J.; Bellemare, M.~G.; Graves, A.; Riedmiller, M.; Fidjeland, A.~K.; Ostrovski, G.; et~al. 2015.
\newblock Human-level control through deep reinforcement learning.
\newblock \emph{nature}, 518(7540): 529--533.

\bibitem[{Neumann and Bartholmai(2015)}]{neumann2015real}
Neumann, P.~P.; and Bartholmai, M. 2015.
\newblock Real-time wind estimation on a micro unmanned aerial vehicle using its inertial measurement unit.
\newblock \emph{Sensors and Actuators A: Physical}, 235: 300--310.

\bibitem[{Neumann et~al.(2013)Neumann, Hernandez~Bennetts, Lilienthal, Bartholmai, and Schiller}]{neumann2013gas}
Neumann, P.~P.; Hernandez~Bennetts, V.; Lilienthal, A.~J.; Bartholmai, M.; and Schiller, J.~H. 2013.
\newblock Gas source localization with a micro-drone using bio-inspired and particle filter-based algorithms.
\newblock \emph{Advanced Robotics}, 27(9): 725--738.

\bibitem[{Parr and Russell(1997)}]{parr1997reinforcement}
Parr, R.; and Russell, S. 1997.
\newblock Reinforcement learning with hierarchies of machines.
\newblock \emph{Advances in neural information processing systems}, 10.

\bibitem[{Pearl(1985)}]{pearl1985bayesian}
Pearl, J. 1985.
\newblock Bayesian netwcrks: A model cf self-activated memory for evidential reasoning.
\newblock In \emph{Proceedings of the 7th conference of the Cognitive Science Society, University of California, Irvine, CA, USA}, 15--17.

\bibitem[{Pearl(2014)}]{pearl2014probabilistic}
Pearl, J. 2014.
\newblock \emph{Probabilistic reasoning in intelligent systems: networks of plausible inference}.
\newblock Elsevier.

\bibitem[{Rhodes, Liu, and Chen(2022)}]{rhodes2022autonomous}
Rhodes, C.; Liu, C.; and Chen, W.-H. 2022.
\newblock Autonomous source term estimation in unknown environments: From a dual control concept to UAV deployment.
\newblock \emph{IEEE Robotics and Automation Letters}, 7(2): 2274--2281.

\bibitem[{Ristic, Arulampalam, and Gordon(2004)}]{Ristic2004BeyondTK}
Ristic, B.; Arulampalam, S.; and Gordon, N.~J. 2004.
\newblock Beyond the Kalman Filter: Particle Filters for Tracking Applications.

\bibitem[{Shi et~al.(2024)Shi, McAreavey, Liu, and Liu}]{shi2024reinforcement}
Shi, Y.; McAreavey, K.; Liu, C.; and Liu, W. 2024.
\newblock Reinforcement Learning for Source Location Estimation: A Multi-Step Approach.
\newblock In \emph{2024 IEEE International Conference on Industrial Technology (ICIT)}, 1--8. IEEE.

\bibitem[{Silver et~al.(2017)Silver, Schrittwieser, Simonyan, Antonoglou, Huang, Guez, Hubert, Baker, Lai, Bolton et~al.}]{silver2017mastering}
Silver, D.; Schrittwieser, J.; Simonyan, K.; Antonoglou, I.; Huang, A.; Guez, A.; Hubert, T.; Baker, L.; Lai, M.; Bolton, A.; et~al. 2017.
\newblock Mastering the game of go without human knowledge.
\newblock \emph{nature}, 550(7676): 354--359.

\bibitem[{Steiner and Bushe(2001)}]{steiner2001large}
Steiner, H.; and Bushe, W. 2001.
\newblock Large eddy simulation of a turbulent reacting jet with conditional source-term estimation.
\newblock \emph{Physics of Fluids}, 13(3): 754--769.

\bibitem[{Tokdar and Kass(2010)}]{tokdar2010importance}
Tokdar, S.~T.; and Kass, R.~E. 2010.
\newblock Importance sampling: a review.
\newblock \emph{Wiley Interdisciplinary Reviews: Computational Statistics}, 2(1): 54--60.

\bibitem[{Vergassola, Villermaux, and Shraiman(2007)}]{vergassola2007infotaxis}
Vergassola, M.; Villermaux, E.; and Shraiman, B.~I. 2007.
\newblock ‘Infotaxis’ as a strategy for searching without gradients.
\newblock \emph{Nature}, 445(7126): 406--409.

\bibitem[{Wang et~al.(2023)Wang, Tian, Wan, Wen, Wang, and Zhang}]{wang2023order}
Wang, X.; Tian, Z.; Wan, Z.; Wen, Y.; Wang, J.; and Zhang, W. 2023.
\newblock Order Matters: Agent-by-agent Policy Optimization.
\newblock In \emph{The Eleventh International Conference on Learning Representations}.

\bibitem[{Wang et~al.(2024)Wang, Zhang, Zhang, Dong, Chen, Wen, and Zhang}]{wang2024zsc}
Wang, X.; Zhang, S.; Zhang, W.; Dong, W.; Chen, J.; Wen, Y.; and Zhang, W. 2024.
\newblock Zsc-eval: An evaluation toolkit and benchmark for multi-agent zero-shot coordination.
\newblock \emph{arXiv preprint arXiv:2310.05208}.

\bibitem[{Wang, Zhang, and Zhang(2022)}]{wang2022model}
Wang, X.; Zhang, Z.; and Zhang, W. 2022.
\newblock Model-based multi-agent reinforcement learning: Recent progress and prospects.
\newblock \emph{arXiv preprint arXiv:2203.10603}.

\bibitem[{Xu et~al.(2023{\natexlab{a}})Xu, Bai, Zhang, Li, and Fan}]{xu2023haven}
Xu, Z.; Bai, Y.; Zhang, B.; Li, D.; and Fan, G. 2023{\natexlab{a}}.
\newblock Haven: Hierarchical cooperative multi-agent reinforcement learning with dual coordination mechanism.
\newblock In \emph{Proceedings of the AAAI Conference on Artificial Intelligence}, volume~37, 11735--11743.

\bibitem[{Xu et~al.(2024)Xu, Mao, Zhang, Xin, Ren, Li, Zhang, Fan, Chen, Wang et~al.}]{xu2024beyond}
Xu, Z.; Mao, H.; Zhang, N.; Xin, X.; Ren, P.; Li, D.; Zhang, B.; Fan, G.; Chen, Z.; Wang, C.; et~al. 2024.
\newblock Beyond Local Views: Global State Inference with Diffusion Models for Cooperative Multi-Agent Reinforcement Learning.
\newblock \emph{arXiv preprint arXiv:2408.09501}.

\bibitem[{Xu et~al.(2023{\natexlab{b}})Xu, Zhang, Li, Zhou, Zhang, and Fan}]{xu2023dual}
Xu, Z.; Zhang, B.; Li, D.; Zhou, G.; Zhang, Z.; and Fan, G. 2023{\natexlab{b}}.
\newblock Dual self-awareness value decomposition framework without individual global max for cooperative multi-agent reinforcement learning.
\newblock \emph{arXiv preprint arXiv:2302.02180}.

\bibitem[{Xu et~al.(2022)Xu, Zhang, Zhan, Baiia, Fan et~al.}]{xu2022mingling}
Xu, Z.; Zhang, B.; Zhan, Y.; Baiia, Y.; Fan, G.; et~al. 2022.
\newblock Mingling foresight with imagination: Model-based cooperative multi-agent reinforcement learning.
\newblock \emph{Advances in Neural Information Processing Systems}, 35: 11327--11340.

\bibitem[{Zadeh(1973)}]{zadeh1973outline}
Zadeh, L.~A. 1973.
\newblock Outline of a new approach to the analysis of complex systems and decision processes.
\newblock \emph{IEEE Transactions on systems, Man, and Cybernetics}, (1): 28--44.

\bibitem[{Zames(1981)}]{zames1981feedback}
Zames, G. 1981.
\newblock Feedback and optimal sensitivity: Model reference transformations, multiplicative seminorms, and approximate inverses.
\newblock \emph{IEEE Transactions on automatic control}, 26(2): 301--320.

\bibitem[{Zhang et~al.(2021)Zhang, Wang, Shen, and Zhou}]{zhang2021model}
Zhang, W.; Wang, X.; Shen, J.; and Zhou, M. 2021.
\newblock Model-based multi-agent policy optimization with adaptive opponent-wise rollouts.
\newblock \emph{arXiv preprint arXiv:2105.03363}.

\bibitem[{Zhao et~al.(2022)Zhao, Chen, Wang, Zhu, Wang, Cheng, Wang, Wang, He, and Liu}]{zhao2022deep}
Zhao, Y.; Chen, B.; Wang, X.; Zhu, Z.; Wang, Y.; Cheng, G.; Wang, R.; Wang, R.; He, M.; and Liu, Y. 2022.
\newblock A deep reinforcement learning based searching method for source localization.
\newblock \emph{Information Sciences}, 588: 67--81.

\bibitem[{Zhao et~al.(2020)Zhao, Chen, Zhu, Chen, Wang, and Ma}]{zhao2020entrotaxis}
Zhao, Y.; Chen, B.; Zhu, Z.; Chen, F.; Wang, Y.; and Ma, D. 2020.
\newblock Entrotaxis-Jump as a hybrid search algorithm for seeking an unknown emission source in a large-scale area with road network constraint.
\newblock \emph{Expert Systems with Applications}, 157: 113484.

\end{thebibliography}
\newpage

\section{Appendix} 
\subsection{Experimental details}
The policy network structure in our experiments consists of a three-layer fully connected neural network, with each layer containing 128 neurons. This architecture is designed to process complex state inputs, effectively extracting relevant features at each layer to inform action selection. Each layer plays a specific role in refining the input information: the first layer captures essential features from raw state data, the second layer further abstracts these features, and the third layer optimizes the representation for precise action decision-making. This structure enables the network to handle a diverse range of reinforcement learning tasks, facilitating the learning of optimal policies by mapping states to action probabilities efficiently. Additional Hyperparameters in experiments are provided as follows:

\begin{table}[h]
\centering
\begin{tabular}{lll}

Hyperparameter & Value & Description  \\ \midrule                              
$lr$ & 1e-4 & The learning rate (lr) of DQN \\
$lr_{actor}$ & 1e-4 & The lr of actor of DDPG      \\
$lr_{actor}$ & 1e-4 & The lr of actor of PPO      \\
$lr_{critic}$ & 1e-5 & The lr of critic of DDPG \\
$lr_{critic}$ & 1e-5 & The lr of critic of PPO \\
Target\_update & 100 & The interval to update network                        \\
 Lambda ($\lambda$) & 0.95 & Balancing variance and bias\\
Minibatch & 64 & The size of train samples\\
Replay buffer & 1000 & The size of replay buffer    \\
discount factor & 0.99 & The discount factor ($\gamma$) \\ \bottomrule                       
\end{tabular}
\caption{Hyperparameter of RL}
\label{Hyperparameter}
\end{table}

\subsection{Related Work}
\textbf{Estimation of environmental dynamics} is crucial in RL, employing a range of techniques to predict how environments respond to agent actions. Traditional methods, like Kalman filtering \cite{kalman1960new,li2015kalman,chui2017kalman} (including EKF: Extended Kalman Filter \cite{anderson2005optimal} and UKF: Unscented Kalman Filter extensions\cite{julier1997new}), particle filtering \cite{doucet2001sequential}, and Bayesian filtering \cite{gordon1993novel}, provide robust state estimation in noisy or uncertain settings. Advanced techniques, such as H$\infty$ filtering \cite{zames1981feedback}, moving window filtering, and model-based RL, further enhance decision-making through predictive models. In recent developments, deep learning methods (e.g., LSTM \cite{graves2012long}) and hybrid physical-data-driven models allow direct learning of complex dynamics from high-dimensional data, supporting adaptability in dynamic environments \cite{li2023passivity}. Complementary methods like system identification, Markov Chain Monte Carlo (MCMC)\cite{metropolis1953equation,geman1984stochastic}, and fuzzy logic enable flexible modeling and state estimation, enhancing automated systems in real-world applications like robotics and climate modeling.

\noindent\textbf{Task progress assessment} employs various techniques to manage and track complex tasks. Knowledge graphs \cite{lassila2001semantic,bollacker2008freebase} help identify dependencies and bottlenecks, while Bayesian networks \cite{pearl1985bayesian,pearl2014probabilistic} provide probabilistic modeling for assessing task success and risks. Fuzzy control \cite{zadeh1973outline}, combined with RL, offers flexible progress estimation in uncertain environments. Hierarchical RL \cite{parr1997reinforcement,kulkarni2016hierarchical} breaks tasks into sub-goals for easier tracking, and goal management dynamically adjusts priorities to align with overarching objectives. Risk-sensitive RL balances progress with risk for high-stakes decisions, and multi-agent task coordination \cite{xu2023dual,xu2023haven,xu2022mingling} enables distributed agents to collaboratively advance sub-tasks. Finally, model-based assessment anticipates delays and obstacles, supporting proactive adjustments to ensure efficient task progression.

\subsection{Normalization Constant in Formula 4}

If we do not omit the normalization constant, the weight update formula will include an explicit normalization step. The complete formula is as follows:

1. Compute the unnormalized weight for each particle:
   \[
   \bar{w}_{k+1}^{i} = w_k^i \cdot Pr(o_{k+1}|\Theta_{k+1}^i)
   \]

2. Compute the sum of the unnormalized weights for all particles:
   \[
   C = \sum_{i=1}^{N} \bar{w}_{k+1}^{i} = \sum_{i=1}^{N} \left( w_k^i \cdot Pr(o_{k+1}|\Theta_{k+1}^i) \right)
   \]

3. Normalize the weight for each particle:
   \[
   w_{k+1}^{i} = \frac{\bar{w}_{k+1}^{i}}{C} = \frac{w_k^i \cdot Pr(o_{k+1}|\Theta_{k+1}^i)}{\sum_{j=1}^{N} \left( w_k^j \cdot Pr(o_{k+1}|\Theta_{k+1}^j) \right)}
   \]

Therefore, the complete weight update formula (without omitting the normalization constant) is:
\[
w_{k+1}^{i} = \frac{w_k^i \cdot Pr(o_{k+1}|\Theta_{k+1}^i)}{\sum_{j=1}^{N} w_k^j \cdot Pr(o_{k+1}|\Theta_{k+1}^j)}
\]

This form clearly shows how each particle's weight is normalized to ensure that the sum of all particle weights equals 1.

\section{Baseline algorithms}
The notations used in these three algorithms (Infotaxis, Entrotaxis, and DCEE) are summarized as follows: \( p_k \) represents the robot's position at time step \( k \), \( u_k \) represents the control input at time step \( k \), \( \hat{s}_k \) represents the estimated source location at time step \( k \), \( \Theta \) represents the vector of unknown parameters, \( Z_k \) represents the set of measurements collected up to time step \( k \), \( b(s) \) represents the belief distribution about the source location given measurements \( Z_k \), \( z_k \) represents the measurement taken at time step \( k \), \( \text{Var}(\hat{s}_{k+1|k}) \) represents the variance of the predicted estimation of the source location at time \( k+1 \), \( H(b(s)) \) represents the entropy of the belief distribution \( b(s) \), \( J(u_k) \) represents the cost function for the respective method, \( f(p_k, u_k) \) represents the state transition function, predicting the next state based on the current state and control input, \( C_{k+1|k} \) represents the predicted covariance matrix of the estimation at time \( k+1 \), and \( \text{trace}(C_{k+1|k}) \) represents the trace of the predicted covariance matrix \( C_{k+1|k} \).

\textbf{Infotaxis}
Infotaxis is an autonomous search strategy where the robot minimizes the predicted variance of the source location estimation. It uses Bayesian inference to update the belief distribution about the source location based on new measurements, continuously refining its understanding of the environment. The core idea is to drive the robot towards actions that maximize the expected information gain, thereby reducing the uncertainty in the source location estimate. This method is particularly useful in environments where the source cannot be directly observed, requiring the robot to infer its location from indirect measurements. Infotaxis has been applied in scenarios such as locating chemical leaks, gas plumes, and even biological searches like finding food sources. This method was proposed by \cite{vergassola2007infotaxis}. in their 2007 paper titled ``‘Infotaxis’ as a strategy for searching without gradients".

\begin{algorithm}
\caption{Infotaxis Algorithm}
\label{alg:Infotaxis}
\begin{algorithmic}[1]
\STATE Initialize robot position $p_k$
\STATE Initialize belief distribution $b(s)$
\WHILE{source not located}
    \STATE Take measurements $z_k$
    \STATE Update belief $b(s) = p(\Theta | Z_k)$ using Bayesian inference
    \STATE Define cost function:
    \STATE $J(u_k) = \min \text{Var}(\hat{s}_{k+1|k})$
    \STATE Select control input $u_k$ that minimizes the cost function:
    \STATE $u_k = \arg\min_{u} J(u_k)$
    \STATE Move robot to new position $p_{k+1} = p_k + u_k$
\ENDWHILE
\end{algorithmic}
\end{algorithm}

\textbf{Entrotaxis}

\begin{algorithm}
\caption{Entrotaxis Algorithm}
\label{alg:Entrotaxis}
\begin{algorithmic}[1]
\STATE Initialize robot position $p_k$
\STATE Initialize belief distribution $b(s)$
\WHILE{source not located}
    \STATE Take measurements $z_k$
    \STATE Update belief $b(s) = p(\Theta | Z_k)$ using Bayesian inference
    \STATE Define entropy of belief distribution:
    \STATE $H(b(s)) = -\sum_i b(s_i) \log b(s_i)$
    \STATE Define cost function:
    \STATE $J(u_k) = \max H(b(s))$
    \STATE Select control input $u_k$ that maximizes the cost function:
    \STATE $u_k = \arg\max_{u} J(u_k)$
    \STATE Move robot to new position $p_{k+1} = p_k + u_k$
\ENDWHILE
\end{algorithmic}
\end{algorithm}
Entrotaxis is an information-theoretic search method that directs the robot to maximize the entropy of the belief distribution. By focusing on maximizing the information gain, the robot is guided to areas with the highest uncertainty in the current belief. This approach leverages the principle that by exploring these high-uncertainty areas, the robot can collect the most informative data, which significantly enhances the accuracy of the source location estimation. Entrotaxis is especially effective in complex and dynamic environments where the source of the signal is obscured or the signal itself is weak and intermittent. Applications include environmental monitoring, search and rescue operations, and hazardous material detection. This method was detailed by \cite{hutchinson2018entrotaxis}. in their 2018 paper "Entrotaxis as a strategy for autonomous search and source reconstruction in turbulent conditions".

\textbf{Dual Control for Exploitation and Exploration (DCEE)}

\begin{algorithm}
\caption{DCEE Algorithm}
\label{alg:DCEE}
\begin{algorithmic}[1]
\STATE Initialize robot position $p_k$
\STATE Initialize belief about source location $\hat{s}_k$
\STATE Initialize particle filter with particles $\{ \Theta_i, w_i \}$
\WHILE{source not located}
    \STATE Take measurements $z_k$
    \STATE Update belief $b(s) = p(\Theta | Z_k)$ using particle filtering:
    \STATE $\{ \Theta_i, w_i \} \leftarrow \text{ParticleFilterUpdate}(\{ \Theta_i, w_i \}, z_k)$
    \STATE Predict next state using state transition function:
    \STATE $p_{k+1|k} = f(p_k, u_k)$
    \STATE Update predicted covariance matrix based on current belief and control input:
    \STATE $C_{k+1|k} = g(b(s), u_k)$
    \STATE Define cost function:
    \STATE $J(u_k) = \| p_{k+1|k} - \hat{s}_{k+1|k} \|^2 + \text{trace}(C_{k+1|k})$
    \STATE Select control input $u_k$ that minimizes the cost function:
    \STATE $u_k = \arg\min_{u} J(u_k)$
    \STATE Move robot to new position $p_{k+1} = p_k + u_k$
\ENDWHILE
\end{algorithmic}
\end{algorithm}

Dual Control for Exploitation and Exploration (DCEE) is a sophisticated control strategy that balances the dual objectives of exploitation (making use of current best estimates) and exploration (actively reducing uncertainty). DCEE employs a particle filter to update the belief distribution about the source location, incorporating measurements taken by the robot. The cost function in DCEE includes both the distance to the estimated source and the uncertainty in this estimate, quantified by the covariance matrix. This dual consideration ensures that the robot's actions not only move it towards the likely location of the source but also strategically probe the environment to gather more data, thus refining future estimates. This balanced approach makes DCEE particularly robust and efficient in scenarios where both the environment and source characteristics are uncertain. Applications of DCEE include robotics for environmental monitoring, industrial inspection, and autonomous exploration in unknown terrains. The DCEE method was discussed in the paper \cite{chen2021dual} "Dual Control for Exploitation and Exploration (DCEE) in autonomous search".

\section{Detailed discussion of the experiment}

\noindent \textbf{Analysis Across Various Particle Numbers}: The success rates for AGDC-enhanced RL algorithms, including AGDC-DQN, AGDC-PPO, and AGDC-DDPG, show a steady increase as the number of particles increases from 100 to 4000. For instance, AGDC-PPO's success rate improves from 88.21\% at 100 particles to 95.27\% at 4000 particles, demonstrating that higher particle numbers enhance the algorithm’s ability to accurately estimate the source by better representing the underlying belief distribution. In contrast, baseline methods like Infotaxis, Entrotaxis, and DCEE also show improved success rates with increased particle numbers, but they consistently lag behind the AGDC-enhanced methods. For example, Infotaxis achieves 88.39\% at 4000 particles, which is significantly lower than AGDC-PPO at the same particle count. The Random method remains ineffective, with success rates consistently below 5\%, as its performance is not influenced by the number of particles. Note: Unlike the other methods that operate in discrete action spaces, DDPG works in a continuous action space, allowing it to move freely within the environment rather than being restricted to discrete steps along predefined edges.

When analyzing the mean traveled distance, AGDC-enhanced RL algorithms demonstrate a consistent reduction as the number of particles increases. For example, AGDC-DDPG reduces its mean traveled distance from 34.33 at 100 particles to 16.83 at 4000 particles, indicating that more particles allow the algorithms to make more informed decisions, leading to shorter and more efficient paths to the source. In contrast, baseline methods, particularly Entrotaxis, exhibit much higher mean traveled distances, even at higher particle numbers. Entrotaxis starts at 148.83 units with 100 particles and only reduces to 78.69 units at 4000 particles, reflecting the inefficiency of these methods in converging to the source, especially with a lower particle count. The Random method, unaffected by particle count, consistently exhibits very high traveled distances, indicating its lack of efficiency.

In terms of search time, AGDC-enhanced RL algorithms consistently outperform the baseline methods by maintaining significantly shorter search times. AGDC-DQN, for instance, keeps a search time of around 0.94 units at 4000 particles, indicating quick convergence to the source. This efficiency can be attributed to the AGDC module's ability to guide the search process more effectively as particle numbers increase. On the other hand, baseline methods such as Infotaxis and Entrotaxis show much longer search times, which continue to increase with higher particle numbers. For instance, Infotaxis requires 3.15 units of time at 4000 particles, indicating slower convergence and less effective search strategies compared to the AGDC-enhanced approaches. The Random method, as expected, has minimal search time but fails to produce meaningful results, further underscoring its ineffectiveness.

\textbf{Analysis Across Various Cessation Thresholds:} The success rates of AGDC-enhanced RL algorithms, including AGDC-DQN, AGDC-PPO, and AGDC-DDPG, improve as the cessation thresholds increase, with AGDC-PPO reaching up to 99.13\% at a threshold of 0.9. This increase in success rate is due to the algorithms accumulating more confidence before stopping, which results in higher accuracy in locating the source. Baseline methods like Infotaxis and Entrotaxis also show improved success rates with higher cessation thresholds, but their performance remains below that of the AGDC-enhanced methods. Although Infotaxis and Entrotaxis can reach around 99\% success rate at the highest thresholds, they require significantly more time and distance to achieve this level of success.

As cessation thresholds increase, AGDC-enhanced RL algorithms demonstrate a decrease in mean traveled distance, indicating more efficient search cessation. For example, AGDC-DDPG reduces its traveled distance to 12.8 units at a threshold of 0.9, reflecting that higher thresholds allow the algorithms to avoid unnecessary exploration and cease the search more efficiently. In contrast, while baseline methods also reduce their traveled distances with higher thresholds, they still require much longer paths compared to AGDC methods. For instance, Infotaxis reduces its traveled distance to 56.4 units at a threshold of 0.9, which is still considerably higher than that of AGDC-DDPG.

Search times for AGDC-enhanced RL algorithms decrease as cessation thresholds increase, with AGDC-PPO reducing its search time to 0.17 units at a threshold of 0.9. This suggests that these algorithms can quickly conclude the search once they have sufficient confidence in their source estimation. However, it is important to note that higher cessation thresholds are not always better. When the threshold exceeds 0.6, the success rate may surpass what would be expected at the upper limit of particle numbers, leading to a misleading improvement in both success rate and distance traveled. This occurs because higher thresholds effectively lower the criteria for task recognition, resulting in the illusion of enhanced performance when in reality, the conditions have simply been made less rigorous.

\section{Descriptions of DQN, PPO, and DDPG}

\subsection{Deep Q-Network (DQN)}

\textbf{Overview}: The Deep Q-Network (DQN) is a reinforcement learning algorithm that combines Q-learning with deep neural networks to approximate the Q-value function. It is designed to handle complex decision-making problems in environments with high-dimensional state spaces.

\textbf{Loss Function}:
The objective of DQN is to minimize the Bellman error of the Q-value function. The loss function is defined as:
\[
L(\theta) = \mathbb{E}_{(s, a, r, s') \sim \mathcal{D}} \left[ \left( r + \gamma \max_{a'} Q(s', a'; \theta^-) - Q(s, a; \theta) \right)^2 \right]
\]
where\\
$\cdot$ \( s \) and \( s' \) are the current and next states, respectively.\\
$\cdot$ \( a \) and \( a' \) are the current and next actions, respectively.\\
$\cdot$ \( r \) is the immediate reward.\\
$\cdot$ \( \gamma \) is the discount factor that controls the importance of future rewards.\\
$\cdot$ \( \theta \) represents the parameters of the Q-network.\\
$\cdot$ \( \theta^- \) represents the parameters of the target network, which is a delayed copy of the Q-network parameters.\\
$\cdot$ \( \mathcal{D} \) is the experience replay buffer.

\noindent\textbf{Key Features}:\\
\textit{Experience Replay}: DQN uses a replay buffer to store past experiences and samples batches randomly during training, which breaks the correlation between samples and improves stability.\\
\textit{Target Network}: The target network is updated less frequently than the Q-network, reducing oscillations and improving training stability.\\
\textit{\(\epsilon\)-Greedy Policy}: DQN uses an \(\epsilon\)-greedy policy that gradually reduces the probability of random exploration, thereby improving decision quality over time.\\

\subsection{Proximal Policy Optimization (PPO)}

\textbf{Overview}: Proximal Policy Optimization (PPO) is an advanced policy gradient method designed to update policies in a stable and efficient manner. PPO achieves this by constraining the magnitude of policy updates, which prevents drastic changes in the policy.

\textbf{Loss Function}:
PPO optimizes the policy by minimizing the following clipped objective function:

\[
L^{\text{CLIP}}(\theta) = \mathbb{E}_t \left[ \min \left( r_t(\theta) \hat{A}_t, \text{clip}\left(r_t(\theta), 1 - \epsilon, 1 + \epsilon\right) \hat{A}_t \right) \right]
\]

\noindent where\\
$\cdot$ \( r_t(\theta) = \frac{\pi_{\theta}(a_t \mid s_t)}{\pi_{\theta_{\text{old}}}(a_t \mid s_t)} \) is the ratio between the new policy and the old policy.\\
$\cdot$ \( \hat{A}_t \) is the advantage estimate, which measures the relative benefit of taking action \( a_t \) in state \( s_t \).\\
$\cdot$ \( \epsilon \) is a hyperparameter that controls the extent of policy updates.

\textbf{Key Features}:
$\cdot$ \textit{Clipped Surrogate Objective:} PPO uses a clipping mechanism to limit the policy update, ensuring that the policy does not change too drastically in a single update, which enhances stability.\\
$\cdot$ \textit{Trust Region Optimization:} PPO effectively keeps the new policy close to the old policy, avoiding instability in the policy update process.\\
$\cdot$ \textit{Mini-batch Updates:} PPO typically updates the policy using small batches of data multiple times, which improves data efficiency and convergence speed.

\subsection{Deep Deterministic Policy Gradient (DDPG)}

\textbf{Overview}: The Deep Deterministic Policy Gradient (DDPG) is a reinforcement learning algorithm designed for continuous action spaces. It combines the advantages of policy gradient methods and Q-learning to achieve efficient decision-making and control in complex environments.

\noindent \textbf{Loss Functions}:
DDPG uses two main loss functions for the Critic and Actor networks:

\noindent \textbf{Critic Loss}:
\[
L(\theta^Q) = \mathbb{E}_{(s, a, r, s') \sim \mathcal{D}} \left[ \left( r + \gamma Q' - Q \right)^2 \right]
\]

\noindent where \( Q(s, a; \theta^Q) \) is the value estimate from the Critic network for the state-action pair \( (s, a) \), \(Q'(s', \mu'(s'; \theta^{\mu'}(Q')) ; \theta^{Q'}) \) is the target value estimate from the target Critic network, \(Q\mu'(s'; \theta^{\mu'}) (Q)\) is the action estimated by the target Actor network.\\

\noindent \textbf{Actor Loss}:

\[
L(\theta^\mu) = -\mathbb{E}_{s \sim \mathcal{D}} \left[ Q(s, \mu(s; \theta^\mu); \theta^Q) \right]
\]

\noindent where \( \mu(s; \theta^\mu) \) is the action selected by the Actor network.

\end{document}